\documentclass[colorinlistoftodos]{article} 
\usepackage{iclr2021_conference,times}

\usepackage{amsmath,amsfonts,bm}

\def\eqref#1{equation~\ref{#1}}

\def\1{\bm{1}}

\DeclareMathAlphabet{\mathsfit}{\encodingdefault}{\sfdefault}{m}{sl}
\SetMathAlphabet{\mathsfit}{bold}{\encodingdefault}{\sfdefault}{bx}{n}

\usepackage{hyperref}
\usepackage{url}
\usepackage{graphicx}
\usepackage{wrapfig}
\usepackage{todonotes}
\usepackage{multirow}
\usepackage{booktabs}
\usepackage{amsmath}
\usepackage{amssymb}
\usepackage{caption}
\usepackage{subcaption}
\usepackage{placeins}

\graphicspath{ {./imgs/} }
\newcommand{\mb}{\mathbf}

\title{Energy-Based Anomaly Detection and Localization}

\author{Ergin Utku Genc, Nilesh Ahuja, Ibrahima J. Ndiour \& Omesh Tickoo\\
Intel Labs\\
\texttt{\{utku.genc,nilesh.ahuja,ibrahima.j.ndiour,omesh.tickoo\}@intel.com}
}

\iclrfinalcopy 
\begin{document}

\maketitle
\begin{abstract}

This brief sketches initial progress towards a unified energy-based solution for the semi-supervised visual anomaly detection and localization problem. In this setup, we have access to only anomaly-free training data and want to detect and identify anomalies of an arbitrary nature on test data. We employ the density estimates from the energy-based model (EBM) as normalcy scores that can be used to discriminate normal images from anomalous ones. Further, we back-propagate the gradients of the energy score with respect to the image in order to generate a gradient map that provides pixel-level spatial localization of the anomalies in the image. In addition to the spatial localization, we show that simple processing of the gradient map can also provide alternative normalcy scores that either match or surpass the detection performance obtained with the energy value. To quantitatively validate the performance of the proposed method, we conduct experiments on the MVTec industrial dataset. Though still preliminary, our results are very promising and reveal the potential of EBMs for simultaneously detecting and localizing unforeseen anomalies in images.
\end{abstract}

\section{Introduction}

\begin{wrapfigure}{l}{0.45\textwidth} 
    \centering
    \includegraphics[width=\linewidth]{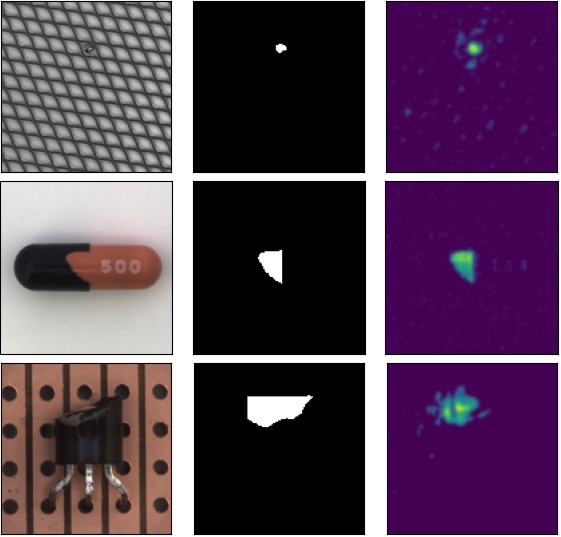}
    \caption{Sample localization outputs for \textit{grid}, \textit{capsule} and \textit{transistor} categories from MVTec AD: From left to right, columns show test image, ground truth mask and predicted localization}
\end{wrapfigure}

The advent of smart manufacturing and the explosion of data collection in factories and other industrial setups have brought considerable attention to the topic of anomaly detection. Its goal is to identify rare, abnormal events from the observation of data. 
This research area has already inspired a vast body of literature whose central idea is to learn a model of normalcy from the positive data and subsequently flag samples that deviate from the learned model as anomalous. Classic approaches include methods based on distances and nearest-neighbor \cite{Knorr_2000,Breunig_2000}, clustering \cite{He03discoveringcluster}, principal component analysis and its kernel variants \cite{Hoffmann_2007}, and one-class support vector machines.
Recently, deep learning has also been used for (deep) anomaly detection, and existing out-of-distribution (OOD) detection methods were successfully applied to this problem \cite{lee2018simple,ahuja2019bdl_dfm,hendrycks2019oe,ren2019likelihood}. In addition, deep generative models such as variational autoencoders (VAE) and generative adversarial networks (GAN) have also been used to characterize abnormality in a variety of manners, based on gradient \cite{9190706}, reconstruction error \cite{Chen_2018}, or density estimation \cite{An2015VariationalAB}. We refer to \cite{Ruff_2021, Goldstein_2016} for an exhaustive review of anomaly detection methods. \linebreak
The overwhelming majority of these methods deal with the detection problem, where one simply needs to discriminate normal samples from anomalous ones. There are numerous usages, medical imaging being one example, wherein it is very important not only to identify abnormal images, but also to localize the abnormality in such images for diagnosis purposes (e.g. areas with brain lesions in CT images).

The literature on anomaly localization is relatively more scant, but includes recent methods based on deep generative models that provide spatial localization by comparing the reconstructed image with the original input \cite{Bergmann_2019,schlegl2017unsupervised,Chen_2018}. While still budding, there is also a new trend of emerging works that focus on anomaly localization \cite{yi_patch_2020,defard2020padim,cohen_sub-image_2021}.

\textbf{Energy-based models for visual anomaly detection and localization.} The principal idea behind energy-based models (EBM) is to learn an \emph{energy function} $E_{\theta}(\mb{x}):\mathbb{R}^D\rightarrow \mathbb{R}$ that parametrizes a density $p_{\theta}(\mb{x})$ as \citep{lecun_tutorial_2006}:
\begin{equation*}
    p_{\theta}(\mb{x}) = \frac{\exp({-E_\theta(\mb{x})})}{\int_\mb{x} \exp({-E_\theta(\mb{x})})},
\end{equation*}
where $\theta$ are parameters of the model. An interesting property of EBMs is that, unlike probabilistic models, they don't necessitate normalization nor a tractable likelihood. Hence, any nonlinear regression function can act as an energy function. These attributes provide great flexibility to EBMs, and we have seen many recent works successfully take advantage of it across different data domains and a variety of energy functions and deep networks. However, being devoid of a tractable likelihood requirement comes with its own challenges. Stable and resource-efficient training of EBMs remains an open problem and recent research has focused on training on popular datasets that were compiled for small-scale classification problems (e.g. CIFAR10). In this regard, the application of EBMs to new problems and different data domains is an opportunity to further develop and popularize EBMs.

Of special importance to this work, EBMs have been shown to have better performance in OOD detection than other likelihood-based generative models \citep{zhai_deep_2016,grathwohl_your_2019,liu_energy-based_2020,du_implicit_2019}. This makes EBMs a promising alternative for semi-supervised anomaly detection problems, where only positive samples (i.e. normal) are available during training. This is especially true for real-world industrial settings where defect data is limited or even unavailable, and the nature of anomalies and defects is not known \emph{a priori}. In those cases, training a model with an explicit discriminative loss term is impractical.

\textbf{Contributions.} This paper proposes a unified EBM method that simultaneously addresses image-level anomaly detection and pixel-level anomaly localization in the semi-supervised setting. Our main contributions can be listed as:
\begin{itemize}
\item An EBM method for visual anomaly detection that jointly tackles detection and localization.
\item The back-propagation of gradients of the EBM energy w.r.t. the image, resulting in a (gradient) heat map from which we can derive both an innovative detection score and pixel-level localization of anomalies in the input image. To our knowledge, this is the first time that gradients of energy score w.r.t input images are interpreted as spatial anomaly information.
\item The application of the method to a real-world industrial case and its quantitative validation with objective metrics. Industrial defect detection stresses the capabilities of generative models as it comes with challenges that can be hardly approximated with synthetic, or research datasets. We provide detailed detection and localization results on the MVTec AD dataset, and compare against reported bechmarks. 
\end{itemize}

\section{Method}
\label{sec:method}

\begin{figure}
\begin{subfigure}{0.45\textwidth}
\includegraphics[width=\textwidth]{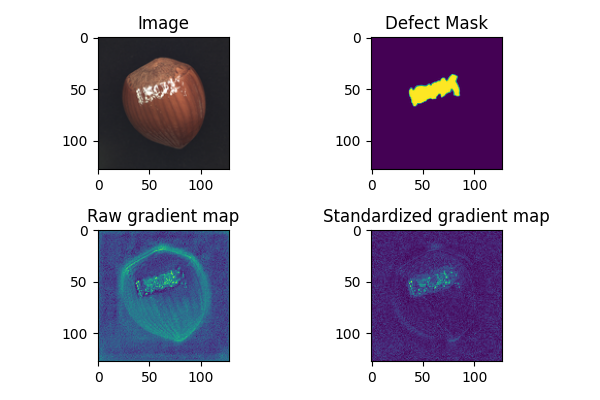}
\caption{Sample image, with visualizations of raw and standardized gradient maps}
\label{fig:raw_vs_std_mask}
\end{subfigure}
\begin{subfigure}{0.45\textwidth}
\includegraphics[width=\textwidth]{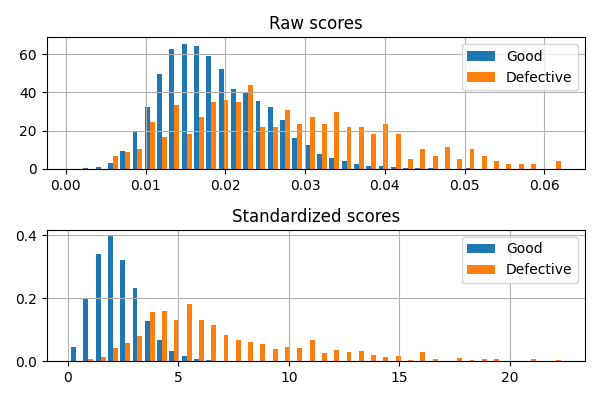}
\caption{Histograms of raw and standardized scores for good vs defective pixels}
\label{fig: raw_vs_std_hist}
\end{subfigure}
\centering
\caption{}
\label{fig:sampleoutput}
\end{figure}

We present the details on our method in this section. Details about the training procedure are provided in Section \ref{sec:experiments}. During inference, an input $\mb{x} \in \mathbb{R}^{w\times h\times c}$ is first processed by the network in a typical forward pass to generate an energy scalar. $w$ and $h$ are the width and height, respectively, of the input, and $c$ is the number of channels ($c=3$ for RGB images, $c=1$ for grayscale images). The gradient of this scalar w.r.t to the input (not the weights) is calculated by back-propagation to produce a gradient-map, $g(\mb{x})$, which has the same shape as that of the input, 
\begin{equation}
    g(\mb{x}) \triangleq \frac{\partial \log p_{\theta}(\mb{x})} {\partial \mb{x}}.
\end{equation}
Let $g_k$ denote the $k$th channel of $g$. A visualization of these gradient-maps shows that although the gradient value is high in the anomalous regions in the image (see Fig. \ref{fig:raw_vs_std_mask}), it is also high in other non-defective regions of the image. Hence,  distinguishing between good and defective pixels based on these raw scores may result in significant false positives. We can significantly improve on this by modeling a probability density $p_{\mb{x}}(g)$ over the gradient-map's values for each pixel location individually. Note that this density is modelled over values of $g$ at a particular pixel location $\mb{x}$. The modeling is performed using the set of gradient-maps, $g^{train}(\mb{x})$, derived from the set of training images. During testing, the log-likelihood score of the observed gradient value, $\hat{g}$, at $\mb{x}$ is calculated as $l=\log p_{\mb{x}}(\hat{g})$ and this score is used as an anomaly score. Hence, the determination of whether a pixel is good or defective is based on how much its gradient value deviates from the typical values observed on the training set images. If the distribution used is a normal distribution, then the log-likelihood, $l$, is equivalent to standardizing the raw anomaly score by its mean, $\mu_k(\mb{x})$, and standard deviation, $\sigma_k(\mb{x})$, as follows 
\begin{equation*}
\l_k(\mb{x}) = \frac{g_k(\mb{x})-\mu_k(\mb{x})}{\sigma_k(\mb{x})}.
\end{equation*}
Note that $\mu_k(\mb{x})$ and $\sigma_k(\mb{x})$ are calculated for each per-pixel location, $\mb{x}$, in each channel $k$ separately, from $g^{train}$ values at $\mb{x}$. The final anomaly score, $a(\mb{x})$ at $\mb{x}$ is the $L_1$ or $L_2$ norm of the corresponding raw or standardized gradient scores from each channel, i.e.
\begin{align}
    a_{raw}(\mb{x}) = \left(\sum_{k=1}^c |g_k(\mb{x})|^r\right)^{1/r}, \quad 
    a_{std}(\mb{x}) = \left(\sum_{k=1}^c |l_k(\mb{x})|^r\right)^{1/r},~r=1,2.
\end{align}
As can be seen in Fig. \ref{fig: raw_vs_std_hist}, standardized scores show much better separation between good and defective pixels. 

Finally, we aggregate the per-pixel anomaly scores to obtain a single score $A(\mb{x})$ for an entire image, again by taking the $L_1$ or $L_2$ norm as follows
\begin{align}
\begin{split}
    A_{raw} &= \|g(\mb{x})\|_r \triangleq \left(\sum_{\mb{x}} |a_{raw}(\mb{x})|^r\right)^{1/r} \\ 
    A_{std} &= \|l(\mb{x})\|_r \triangleq \left(\sum_{\mb{x}} |a_{std}(\mb{x})|^r\right)^{1/r} 
\end{split}
\end{align}
This score can be used to for the classification of the image as defective or good. Note that $A_{raw}$ is essentially the same as approximate mass score proposed in \citep{grathwohl_your_2019}.

\section{Experiments}
\label{sec:experiments}
\subsection{Training}
In this work, we take an MCMC-based maximum likelihood (ML) approach to training our model. As our energy function approximator, we use an all convolutional neural network similar to the one \citet{nijkamp_learning_2019} used (see Table~\ref{tab:cnn-table} in the Appendix), but instead of \emph{Leaky-ReLU}, we use \emph{ELU} \citep{clevert_fast_2016} activation functions between all layers to stabilize training. To optimize for likelihood, we use Contrastive Divergence (CD) \citep{hinton_training_2002}.

To synthesize the negative samples needed by the CD objective, we employ a Stochastic Gradient Langevin Dynamics (SGLD) MCMC sampler \citep{welling_bayesian_2011}. We initialize MCMC chains from a \textit{uniform} distribution and run a fixed number of MCMC steps ($100$). For a more resource-efficient training, we experimented with reducing the number of steps and using a \emph{replay buffer} to persist MCMC chains \citep{tieleman_training_2008} instead of initalizing from noise at each training iteration, but we could not manage to find a stable training regime on all categories and had sporadic energy spikes. We leave the solution of these problems to later work and share only the results of \emph{non-persistent} trainings for consistency over all dataset categories. During sampling, we don't accumulate gradients w.r.t. model parameters for later use at model parameter update step.

During training, we only use nondefective images and we don't explicitly optimize for any discriminative objective. In that sense, training is unsupervised. But, as we know that the training data is not contaminated with defective samples, we prefer to call our approach \emph{semi-supervised}. We believe that a fully unsupervised method should tackle the consequences of possible training data contamination.

\subsection{Results}

\begin{table}
    \footnotesize
    \captionsetup{width=0.75\textwidth}
	\caption{AUROC results for anomaly detection on MVTec (results for other benchmarks taken from \cite{cohen_sub-image_2021})}
	\label{table:detection_roc}
	\centering
	\begin{tabular}{ccccccc}
		\toprule
		\midrule
		& Geom & GANomaly & AE(L2) & ITAE & SPADE & EBM (Ours) \\
		\midrule
		Average &  0.67 & 0.76 & 0.75 & 0.84 & \textbf{0.85} & 0.72 \\
		\bottomrule
	\end{tabular}
\end{table}

\begin{figure}[h!]
\centering
\includegraphics[width=0.96\linewidth]{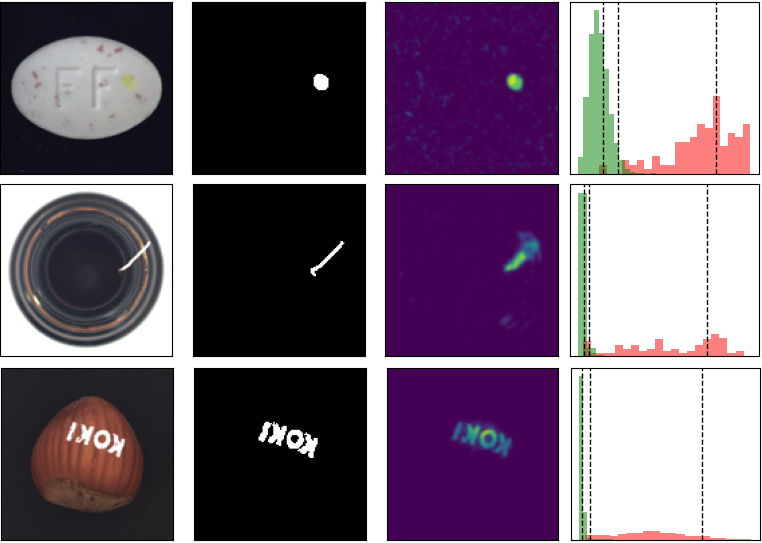}
\includegraphics[width=0.96\linewidth]{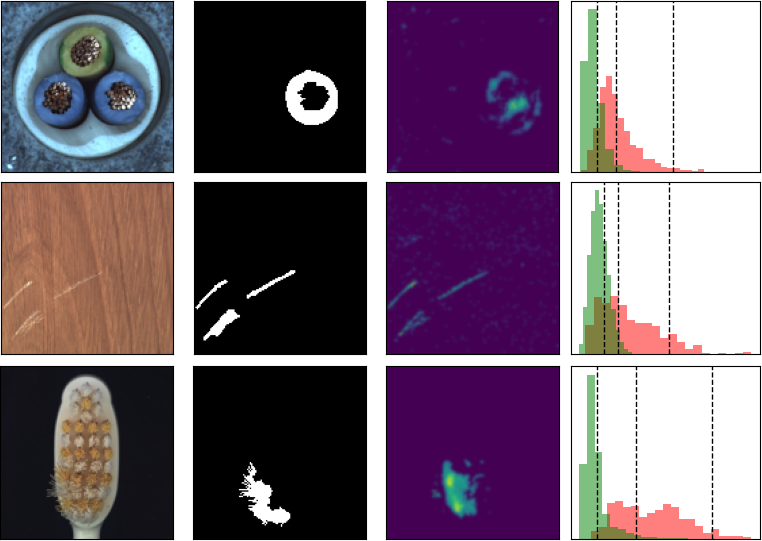}
    \caption{From left to right: sample images with anomalies, their corresponding ground truth masks, anomaly score maps (standardized gradient maps), and histograms (good vs. defective pixels). Vertical dashed lines are \emph{three-sigma rule} thresholds.}
\label{fig:mask_images}
\end{figure}

\begin{table}
    \footnotesize
    \captionsetup{width=0.75\textwidth}
	\caption{AUROC results for anomaly localization on MVTec (results for other benchmarks taken from \cite{Bergmann_2019})}
	\label{table:localization_roc}
	\centering
	\begin{tabular}{cccccc}
		\toprule
		\midrule
		 Category & \multirow{2}{*}{\begin{tabular}{c}
 				AE  \\ (SSIM)
 		\end{tabular}} & \multirow{2}{*}{\begin{tabular}{c}
 				AE  \\ (L2)
 		\end{tabular}} & AnoGAN & \multirow{3}{*}{\begin{tabular}{c}
 				CNN  \\ Feature \\ Dictionary
 		\end{tabular}} & \multirow{2}{*}{\begin{tabular}{c}
 				EBM  \\ (Ours)
 		\end{tabular}} \\
 		\\
 		\\
		\midrule
    Carpet & \textbf{0.87} & 0.59 & 0.54 & 0.72 & 0.63 \\
		\midrule
    Grid & \textbf{0.94} & 0.90 & 0.58 & 0.59 & 0.86 \\
		\midrule
    Leather & 0.78 & 0.75 & 0.64 & \textbf{0.87} & \textbf{0.87}\\
		\midrule
    Tile & 0.59 & 0.51 & 0.50 & \textbf{0.93} &  0.57\\
		\midrule
    Wood & 0.73 & 0.73 & 0.62 & \textbf{0.91} & 0.74\\
		\midrule
    Bottle & \textbf{0.93} & 0.86 & 0.86 & 0.78 & 0.72\\
		\midrule
    Cable & 0.82 & \textbf{0.86} & 0.78 & 0.79 & 0.56\\
		\midrule
    Capsule & \textbf{0.94} & 0.88 & 0.84 & 0.84 & 0.64\\
		\midrule
    Hazelnut & \textbf{0.97} & 0.95 & 0.87 & 0.72 & 0.78\\
		\midrule
    Metal Nut & \textbf{0.89} & 0.86 & 0.76 & 0.82 & 0.65\\
		\midrule
    Pill & \textbf{0.91} & 0.85 & 0.87 & 0.68 & 0.75\\
		\midrule
    Screw & \textbf{0.96} & \textbf{0.96} & 0.80 & 0.87 & 0.87 \\
		\midrule
    Toothbrush & 0.92 & \textbf{0.93} & 0.90 & 0.77 & 0.68\\
		\midrule
    Transistor & \textbf{0.90} & 0.86 & 0.80 & 0.66 & 0.74\\
		\midrule
    Zipper & \textbf{0.88} & 0.77 & 0.78 & 0.76 & 0.55\\		
		\bottomrule
	\end{tabular}
\end{table}

We test our approach on the problem of visual anomaly detection on the recently introduced MVTec dataset \citep{bergmann_mvtec_2019}, which is a dataset for benchmarking anomaly detection methods with a focus on industrial inspection. There are fifteen different object and texture categories in the dataset, and we train a separate EBM for each category. The training is performed using defect-free training images only and the model is not exposed to any defective images during training. Training of EBMs is known to be computationally demanding and hence the input images are downsized, to $128\times 128$ to keep the training time manageable.

For a test input, we generate both the per-pixel anomaly scores, $a\mb{(x)}$, and the per-frame anomaly scores, $A(\mb{x})$ as described in Section \ref{sec:method}. These scores can be used to distinguish between good and defective data, at both the pixel level and the frame level. This effectively creates a binary classifier, whose performance is characterized by the receiver operating characteristics (ROC) curve. We tested performance with both the raw and standardized scores, and found that the standardized scores generally outperform the raw scores and hence, we report results using standardized scores. For the detection task, we report the area under the ROC curve (AUROC) in Table \ref{table:detection_roc}, and for the localization task, we report the AUROC in Table \ref{table:localization_roc}. We can observe that the EBM achieves respectable scores on both tasks, but there is certainly room for improvement. For a comparison between raw and standardized scores, please refer to section \ref{sec:rvs} in the Appendix. For most categories, it is observed that standardizing the scores results in an improvement in the performance. 
For localization, sample images containing anomalies are shown in Fig. \ref{fig:mask_images}, along with the standardized gradient maps produced by our method. For this set, we see that the high-intensity regions in the anomaly maps correspond well visually with ground truths. The accompanying density histograms show good separation of scores between good-pixels and anomalous pixels.

\section{Conclusion}

This work presented an energy-based method for the semi-supervised anomaly detection and localization problem. We propose an alternative scoring function for image-level anomaly detection and spatial anomaly maps for pixel-level localization; both are derived from back-propagating gradients of the energy score w.r.t. the test image. We experiment our approach on the MVTec AD industrial dataset and benchmark against the state of the art. While the method is still in the early stages of development, initial results are encouraging with indications of how to obtain further improvements. These include using larger models with higher capacity to model the energy function, and exploring regimens to perform training in a more stable and less resource-intensive manner.


\bibliography{iclr2021_conference}
\bibliographystyle{iclr2021_conference}

\newpage
\appendix
\section{Appendix}



\subsection{CNN Topology}
\begin{table}[h!]
\caption{CNN topology}
\label{tab:cnn-table}
\begin{center}
\begin{tabular}{cccccc}
\toprule
\bf kernel & \bf stride & \bf padding & \bf $f_{in}$ & \bf $f_{out}$ & \bf activation \\
\midrule
(3x3) & 1 & 1 & $n_c={1,3}$ & 32 & ELU \\
(4x4) & 2 & 1 & 32 & 64 & ELU \\
(4x4) & 2 & 1 & 64 & 128 & ELU \\
(4x4) & 2 & 1 & 128 & 256 & ELU \\
(4x4) & 2 & 1 & 256 & 256 & ELU \\
(4x4) & 2 & 1 & 256 & 256 & ELU \\
(4x4) & 1 & 0 & 256 & 1 & N/A \\
\bottomrule
\end{tabular}
\end{center}
\end{table}

\subsection{Raw vs Standardized scores}
\label{sec:rvs}
We compare AUROC scores for both detection and localization using raw gradient-map scores and standardized scores. In general, we observe that standardized scores perform better than raw scores. For a few categories, the raw scores do provide better AUROC.

\begin{table}[h]
    \footnotesize
	\caption{Comparing Raw and Standardized AUROC scores for detection}
	\label{table:raw_vs_std_detect}
	\centering
	\begin{tabular}{cccc}
		\toprule
		& Energy & Raw gradients & Standardized gradients \\
		\midrule
		Average & 0.56 & 0.69 & 0.72 \\
		\bottomrule
	\end{tabular}
\end{table}

\begin{table}[h]
    \footnotesize
	\caption{Comparing Raw and Standardized AUROC scores for localization}
	\label{table:raw_vs_std_loc}
	\centering
	\begin{tabular}{cccc}
		\toprule
 		Category & Raw & Standardized & Difference \\
		 
		\midrule
    Carpet & 0.53 & 0.63 & 0.10 \\
		\midrule
    Grid & 0.86 & 0.86 & 0.00 \\
		\midrule
    Leather & 0.43 & 0.86 & 0.43\\
		\midrule
    Tile & 0.50 & 0.57 & 0.07\\
		\midrule
    Wood & 0.73 & 0.74 & 0.01\\
		\midrule
    Bottle & 0.70 & 0.72 & 0.02\\
		\midrule
    Cable & 0.46 & 0.56 & 0.10\\
		\midrule
    Capsule & 0.44 & 0.64 & 0.20\\
		\midrule
    Hazelnut & 0.73 & 0.78 & 0.05\\
		\midrule
    Metal Nut & 0.71 & 0.65 & -0.06\\
		\midrule
    Pill & 0.71 & 0.74 & 0.03\\
		\midrule
    Screw & 0.88 & 0.87 & -0.01 \\
		\midrule
    Toothbrush & 0.82 & 0.68 & -0.14\\
		\midrule
    Transistor & 0.74 & 0.74 & 0.00\\
		\midrule
    Zipper & 0.64 & 0.55 & -0.09\\		
		\bottomrule
	\end{tabular}
\end{table}

\end{document}